\documentclass[runningheads]{llncs}
\usepackage{graphicx}
\usepackage{comment}
\usepackage{amsmath,amssymb} 
\usepackage{color}

\usepackage[linesnumbered,boxed,ruled,commentsnumbered]{algorithm2e}
\usepackage{booktabs}
\usepackage{threeparttable}
\usepackage{multirow}
\usepackage{array}

\usepackage[width=122mm,left=12mm,paperwidth=146mm,height=193mm,top=12mm,paperheight=217mm]{geometry}

\begin{document}
\pagestyle{headings}
\mainmatter
\title{Learning Multi-modal Information for Robust Light Field Depth Estimation } 
\author{Yongri Piao, Xinxin Ji, Miao Zhang\thanks{Corresponding Author},Yukun Zhang}
\institute{ Dalian University of Technology, China\\
        {\tt\small\{yrpiao, miaozhang\}@dlut.edu.cn}
        {\tt\small\{jxx0709,zhangyukun\}@mail.dlut.edu.cn}}

\maketitle
\begin{abstract}
\emph{\emph{Light field data has been demonstrated to facilitate the depth estimation task. Most learning-based methods estimate the depth information from EPI or sub-aperture images, while less methods pay attention to the focal stack. Existing learning-based depth estimation methods from the focal stack lead to suboptimal performance because of the defocus blur. In this paper, we propose a multi-modal learning method for robust light field depth estimation. We first excavate the internal spatial correlation by designing a context reasoning unit which separately extracts comprehensive contextual information from the focal stack and RGB images. Then we integrate the contextual information by exploiting a attention-guide cross-modal fusion module. Extensive experiments demonstrate that our method achieves superior performance than existing representative methods on two light field datasets. Moreover, visual results on a mobile phone dataset show that our method can be widely used in daily life.Codes are available: \textcolor{blue}{https://github.com/OIPLab-DUT/Deep-Light-Field-Depth-Estimation}}}
\end{abstract}

\section{Introduction}

The 4D light field camera can simultaneously record spatial and angular information of light rays incident at pixels of the tensor by inserting a microlens array between the main lens and image sensor. Due to its abundant information captured through one imaging, compared to traditional 2D images, the additional angular information can help to synthesize focal stacks and all-focus images with the rendering \cite{levoy1996light} and post-capture refocusing technique \cite{ng2005light}. These advantages make the light field data has been successfully applied to depth estimation.

Light field depth estimation has become a popular research topic and plays a more and more important role in a wide range of applications, such as scene reconstruction \cite{kim2013scene}, image super-resolution \cite{pujades2014bayesian,wanner2013variational}, object tracking \cite{yang2013new}, saliency detection \cite{piao2019deep}, image segmentation \cite{zhu20174d}.
According to the input data type, existing depth estimation methods from light field can be categorized into three types: depth estimation based on  epipolar plane images (EPIs) \cite{kim2014cost,heber2017neural}, depth estimation based on sub-aperture images (sub-apertures) \cite{tomioka2017depth,wang2016depth} and depth estimation based on the focal stack \cite{pertuz2013analysis,tao2013depth,lin2015depth}. The EPIs-based depth estimation explores geometry structures in EPIs to capture the depth information. The sub-apertures-based depth estimation usually predicts depth maps by regarding sub-aperture images as a multi-view stereo configuration. The depth estimation based on focal stack often estimates the depth by utilizing some cues of the focal stack such as defocus cue, shading cue, symmetry, etc. Despite the above three types of the traditional methods have achieved great success, there are still some challenges that limit their application. For example, because of depending on domain-specific prior knowledge to increase the robustness, the generalization ability in different scenarios is limited to get the unsatisfactory depth map. With the development of deep learning, learning-based light field depth estimation emerges and alleviates this problem. Many methods based on deep learning capture depth cues from EPIs or sub-aperture images, while less methods focus on the focal stack.

Based on this observation, we plan to predict depth maps with the focal stack in this paper. Focal stack generated from light field contains the focusness information which can help focus at the object in different range of depth. However, focal slices only contain the local focusness information of a scene. In other words, the defocus information may decrease the accuracy of the prediction results. Therefore, it is not enough to get a more robust depth by only using the focal stack \cite{hazirbas2018deep}. Considering the RGB image contains global and high quality structure information, we have strong reasons to believe that incorporating the focal stack and RGB images is helpful for light field depth estimation.

In order to improving the performance of depth estimation, there are still some issues needed to be consider. First, we need to focus on the fact that depth value of each pixel is related to the neighboring pixel. Therefore, it is difficult to predict the accurate depth value when considered in isolation, as local image evidence is ambiguous. So, how to effectively capture contextual information to find the long-range correlations between features is essential for reasoning small and thin objects and modeling object co-occurrences in a scene. Second, since the RGB image contains more internal details and the focal slices contain abundant depth information, how to effectively capture and integrate the complementary information between them is an important aspect we should concentrate on.

In this paper, our method confronts these challenges successfully. In summary, our main contributions are as follows:
\begin{itemize}
\item we propose a graph convolution-based context reasoning unit (CRU) to comprehensively extract contextual information of the focal stack and RGB images, respectively. Such a design allows to explore the internal spatial correlation between different objects and regions. It's helpful to clarify local depth confusions and improve the accuracy of depth estimation.

\item we propose a attention-guided cross-modal fusion module (CMFA) to integrate different information in the focal stack and RGB images. It captures the complementary information from paired focal slices and RGB features by cross-residual connections to enhance features, and then learns multiple attention weights to integrate the multi-modal information effectively. It's helpful to compensate for the detail loss caused by defocus blur.

\item Extensive experiments on two light field datasets show that our method achieves consistently superior performance over state-of-the-art approaches. Moreover, our method is successfully applied to the dataset collected by mobile phones. This demonstrates our method is not limited to the focal stack and is more practical for daily life.
 \end{itemize}
\section{Related Work}

{\bfseries{Traditional methods. }}For the depth estimation from light field images, traditional methods make use of the sub-aperture images or epipolar plane images (EPIs) or focal stacks. In terms of sub-aperture images, Georgiev and Lumsdaine \cite{georgiev2010reducing} compute a normalized cross correlation between microlens images to estimate disparity maps; Bishop and Favaro \cite{bishop2011light} propose a interactive method for a multi-view stereo image; Yu \emph{et al}. \cite{yu2013line} analyze the 3D geometry of lines and compute the disparity maps through line matching between the sub-aperture images; Heber and Pock \cite{heber2014shape} use the low-rank structure regularization to align the sub-aperture images for estimating disparity maps; Jeon \emph{et al}. \cite{jeon2015accurate} use the cost volume to estimate the multi-view stereo correspondences with sub-pixels. The early works about EPIs can be traced back to the research by Bolles \emph{et al}. \cite{bolles1987epipolar} who estimate the 3D structure by detecting edges in EPIs and fitting straight-line segments to edges afterwards. Zhang \emph{et al}. \cite{zhang2016robust} propose a local depth estimation method which employs the matching lines and spinning parallelogram operator to remove the effect of occlusion; Zhang \emph{et al}. \cite{zhang2016light} exploit the linear structure of EPIs and locally linear embedding to predict the depth map; Johannsen \emph{et al}. \cite{johannsen2016sparse} employ a specially designed sparse decomposition which leverages the orientation depth relationship on its EPIs; Wanner and Goldluecke \cite{wanner2013variational} compute the vertical and horizontal slopes of EPIs using a structured tensor, formulate the depth map estimation problem as a global optimization approach and refine initial disparity maps using a fast total variation denoising filter; Sheng \emph{et al}. \cite{sheng2018occlusion} propose a method which combines the local depth with occlusion orientation and employs multi-orientation EPIs. Tao \emph{et al}. \cite{tao2013depth} exploit both defocus cues and correspondence \emph{}cues from the focal stack to achieve better performance by using a contrast based measure. However there is a problem with these methods, which are usually too dependent on prior knowledge to generalize to other datasets easily.

\noindent{{\bfseries{Learning-based Methods. }}}Recently, convolutional neural networks have performed very well on computer vision tasks such as segmentation \cite{wang2019fast} and classification \cite{kipf2016semi}. However, there are fewer learning-based methods for depth estimation from light field images. Heber \emph{et al}. \cite{heber2016convolutional} propose a network which consists of encoder and decoder parts to predict EPI line orientations. Luo \emph{et al}. \cite{luo2017epi} propose an EPI-patch based CNN architecture and Zhou \emph{et al}. \cite{zhou2018scale} introduce a scale and orientation aware EPI-Patch learning network. Shin \emph{et al}. \cite{shin2018epinet} propose a fully CNN for depth estimation by using the light field geometry. Anwar \emph{et al}. \cite{anwar2017depth} exploit dense overlapping patches to predict depth from a single focal slice. Hazirbas\emph{ et al}. \cite{hazirbas2018deep} propose the first method based deep-learning to compute depth from focal stack. These methods have greatly improve the prediction results but there is still room for improvement. Moreover, these methods pay less attention to the focal stack and the corresponding RGB images. In this paper, we propose a deep learning-based method which effectively incorporates a RGB and a focal stack to improve the prediction. 

\section{Method}
In this section, we focus on the problem how to make effective use of the RGB image and focal stack to predict depth maps. First, we briefly introduce the overall architecture which can be trained end-to-end in Sec.3.1. Then we detail our context reasoning unit (CRU) and its key component in Sec.3.2. Finally, we elaborate on the attention-guided cross-modal fusion module (CMFA) which effectively captures and integrates the paired focal slices and RGB features to significantly improve the performance in Sec.3.3.

  \begin{figure*}[!ht]
\begin{center}
\includegraphics[width=0.9\linewidth]{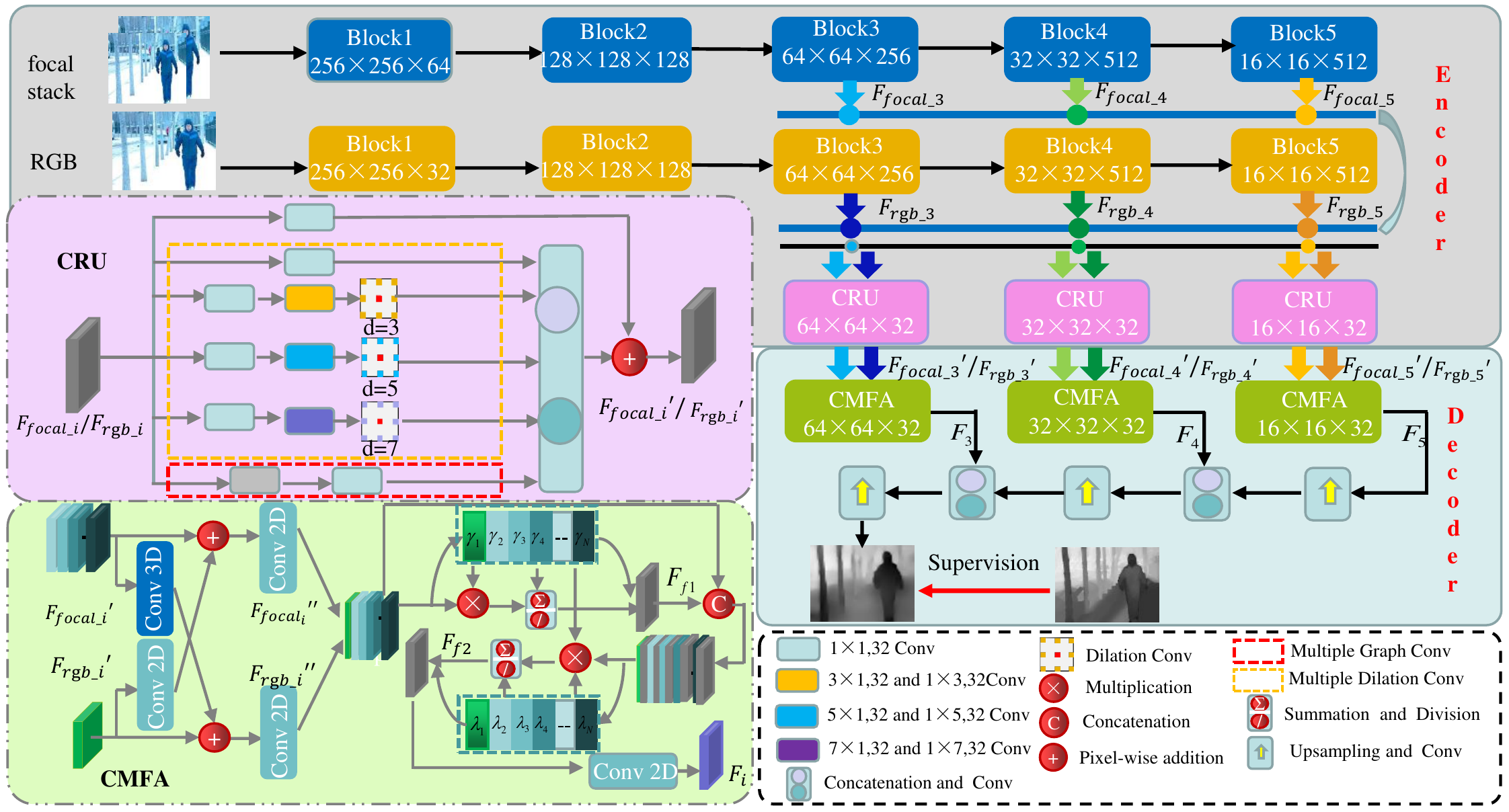}
\end{center}
\vspace{-4mm}
   \caption{The whole pipeline of our method. It consists of the encoder and decoder.  }
\label{fig:long}
\vspace{-6mm}
\end{figure*}

\subsection{The Overall Architecture}

Our network architecture consists of an encoder and a decoder. It aims to comprehensively extract and effectively integrate features from the focal stack and the RGB image. The overall framework is shown in Fig.1. The encoder has a symmetric two-streams feature extraction structure: focal stack stream and RGB stream. Each stream contains the backbone achieved by VGG-16 \cite{simonyan2014very} that last pooling and fully-connected layers are discarded, and the context reasoning unit (CRU). The decoder consists of a progressive hierarchical fusion structure which contains our proposed attention-guided cross-modal fusion module (CMFA).

For the sake of explanation, the shape of features can be denoted as $N\times W\times H\times C $. N, W, H and C represent the number of focal slices, height, width and channel respectively. Specifically, given the RGB image $I_0$ and the focal stack consisted of 12 focal slices $\{I_1,I_2,...I_{12}\}$, we separately feed them into the backbones. Then, with the raw side-out light field features $\{{F_{focal\_i} }\}_{i = 3}^5$ ($12\times W\times H\times C$) and RGB features $\{{F_{rgb\_i} }\}_{i = 3}^5$ ($1\times W\times H\times C$) from the last three layers, we respectively utilize the CRU to comprehensively extract contextual information from them. It can find the long-range correlations between features to explore the internal spatial correlation of focal slices and RGB images. In the decoder, we boost our model by the proposed attention-guided cross-modal fusion module (CMFA) which integrates paired feature $\{{{F_{focal\_i} }}'\}_{i = 3}^5$ ($12\times W_{1}\times H_{1}\times C_{1}$) and $\{{{F_{rgb\_i} }}'\}_{i = 3}^5$ ($1\times W_{1}\times H_{1}\times C_{1}$) which are come from the CRU to get $\{{F_i }\}_{i = 3}^5$ ($1\times W_{2}\times H_{2}\times C_{2}$). Finally, the multi-level features are decoded by the top-down architecture and depth maps are supervised by the ground truths.

\subsection{Context Reasoning Unit (CRU)}

As focus and refocus regions of focal slices indicate the different depth information and the RGB image contains more detail information of a scene, it is important to extract contextual information for capturing long-range correlation between features and exploring internal spatial correlation in scenes. To do this, we propose a context reasoning unit (CRU). Different from the general context aggregation modules which use the pure ASPP \cite{DBLP:journals/corr/ChenPK0Y16} or combine the ASPP with image-level encoder, our CRU can not only capture the spatial correlations between large objects with multiple dilated convolutions, but also pay more attention to the small and thin objects by capturing more abstract features in the image with multiple graph convolutions.

As illustrated in Fig.1, the CRU consists of three branches. The top one is a short-connection operation which learns the residual information, the middle branch is multiple dilated convolutions and the bottom branch is the multiple graph convolutions. In the unit, the output features from the middle branch and the bottom branch are concatenated and convolved, and then added to the features from the top branch to get the final refined features.

The multiple dilated convolutions consist of a cross-channel learner and an atrous spatial pyramid pooling. They can learn complex cross-channel interactions by a $1\times1$ convolution operation, and extract features via dilated convolution with small rates as 3, 5, 7. So it captures multi-scale spatial information. The resulting features are dominated by large objects, which effectively model the spatial correlations between the large objects in the image.

\begin{figure}[!ht]
\begin{center}
\includegraphics[width=0.9\linewidth]{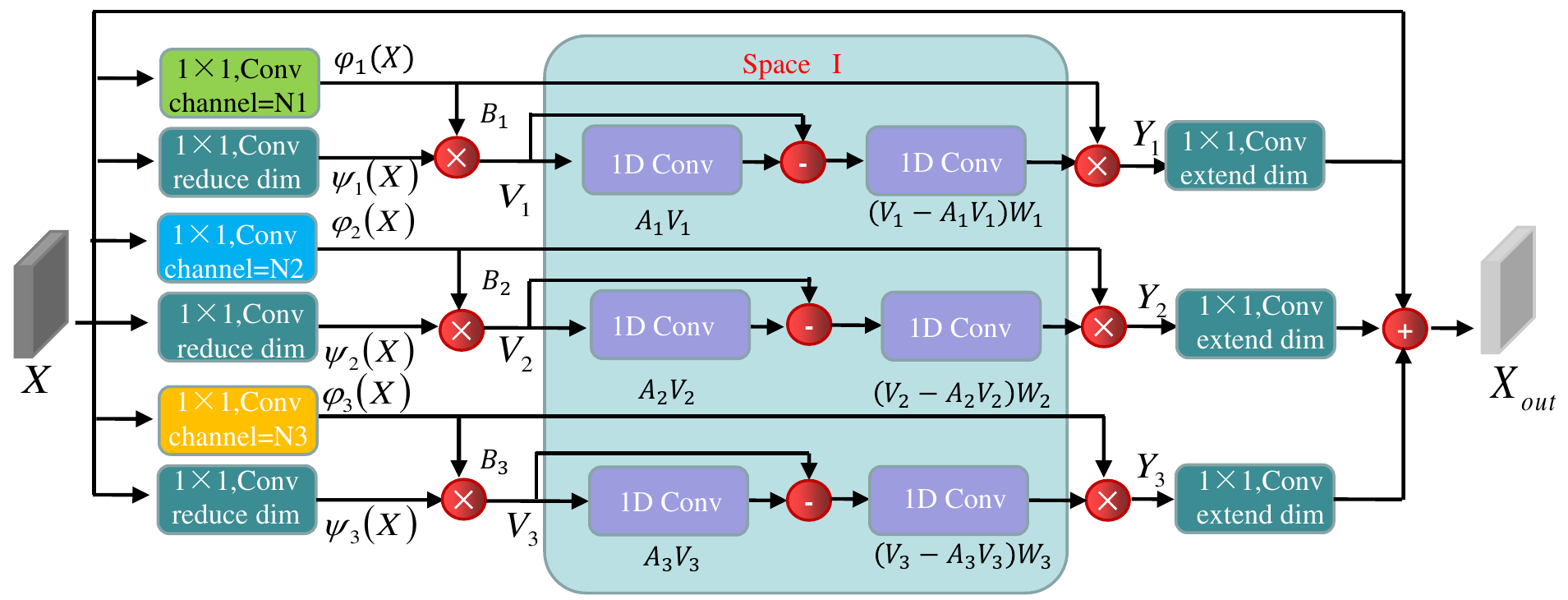}
\end{center}
\vspace{-4mm}
   \caption{ The architecture of multiple graph convolutions.}
\label{fig:short1}
\vspace{-6mm}
\end{figure}

The multiple graph convolutions model interdependencies along the channel dimensions of the network's feature map. Different from previous methods which adopt image-level encoder structures, such as pure \emph{fc} layers \cite{eigen2014depth} or global average pooling \cite{hu2018squeeze}, our design can effectively model and communicate information from the region-level clues with less parameters and more nodes. Compared to \cite{chen2019graph}, we establish multiple node topological graphs on parallel to cover regions at different scales. The number of nodes in graphs are changing dynamically according to the spatial size of input features. These enable our network to refine spatial relationship between different regions and adapt to the small and thin objects effectively. Therefore, we can produce coherent predictions that consider all objects and regional differences in the image. Fig.2 shows the schematic illustration of our design.

Specifically, take the raw side-out focal features $F_{focal\_i}$ ($ 12\times W\times H\times C$) as an example. Given the input features $X=F_{focal\_i}$, we establish three node topological graphs through three parallel branches to refine the spatial relationship. In the \emph{i}-th branch (\emph{i} = 1, 2, 3), the process can be divided to three steps: 1) Space projection: mapping the features from Coordinate Space \emph{S} to Interaction Space \emph{I}. We first reduce the dimension of $X$ with $\psi _i(X)$ and formulate the projection function $\varphi_i(X)=B_i$. In practice,  $\psi _i(X)$ is achieved by a $1\times1$ convolution layer with $C_i$ channels and $\varphi _i(X)$ is achieved by a $1\times1$ convolution layer with $ N_i=\frac{W \times H}{{4 \times 2^{i - 1} }}$ channels. Therefore, the input features $X$ is projected to a new feature $V_i$ ($12 \times N_i \times C_i$) in the space \emph{I}. $V_i$ integrates information from different regions of the focal slices. Note the $N_i$ is the number of nodes and it is are changing dynamically according to the spatial size of raw features. This design helps our network effectively adapt to the features with different scales. 2) Feature graph convolution: reasoning the relation with graphs. After projection, we can build a fully-connected graph with adjacency matrix $A_i$ ($ 12 \times N_i \times N_i $) in the Interaction Space \emph{I}, where each node contains the feature descriptor. Therefore, the context reasoning problem is simplified to interaction capturing between nodes. They are achieved by two 1D convolution layers along the channel and node directions. With the adjacency matrix and layer-specific trainable edge weights $W_i$, we can diffuse information across nodes to get the node-feature matrix $M_i$. 3) Reprojection: mapping the features to Coordinate Space \emph{S} from Interaction Space \emph{I}. After the reasoning, we map the new features $M_i$ back into the original coordinate space with the another mapping function ${B_i}^T$ to get $Y_i$ ($ 12\times W \times H\times C_i$). $Y_i$ is extended to the original dimension by a $1\times1$ convolutional layer. Finally, we add the output features $\{ {Y_i } \}_{i = 1}^3$ to the original features $X$. Another convolution layer is attached for the final feature $X_{out}$ ($ 12\times W\times H\times C$). The implementation process can be defined as follows:
\begin{equation}
\begin{array}{l}
\{ {V_i }\}_{i = 1}^3  =\{ {B_i \psi _i }\}_{i = 1}^3 = \{ \varphi_i(X) \psi _i(X)\}_{j = 1}^3
 \end{array}
\end{equation}
\begin{equation}
\begin{array}{l}
\{ {M_i } \}_{i = 1}^3  = (V_i  - A_{i} V_i )W_i
 \end{array}
\end{equation}
\begin{equation}
\begin{array}{l}
\{ {Y_i }\}_{i = 1}^3  = \{ {(B_i )^T M_i } \}_{i = 1}^3
 \end{array}
\end{equation}
\begin{equation}
\begin{array}{l}
 X_{out}  = X + Y_1 + Y_2 + Y_3
 \end{array},
\end{equation}
For the RGB images, the same operations are performed on the RGB features generated from the backbone.

\subsection{Attention-guided Cross-Modal Fusion module (CMFA)}
The defocus blur could lead to the detail loss which negatively affects the accuracy of the depth map. To address this challenge, with the contextual information extracted from focal slices features and RGB features by the context reasoning unit (CRU), we aim to fuse them to compensate for detail loss. The direct method is to simply concatenating the RGB features and the focal slices features, but this not only ignores the relative contribution of different focal slices features and RGB features to the final result, but also severely destroys the spatial correlation between focal slices. Therefore, we design a attention-guide cross-modal fusion module to integrate the implicit depth information in focal slices and the abundant content information in RGB images effectively.

As shown in Fig.1, the process of our module can be divided to two steps: 1) capturing complementary information to enhance features; 2) fusing the enhanced features. Considering that the difference between the focal features and RGB features, we first augment them to highlight the complementarity of multi-modal information. In the first step, we introduce the cross-modal residual connections achieved by simple 3D convolution and 2D convolution to capture complementary information from paired features $\{{{F_{focal\_i} }}'\}_{i = 3}^5$ and $\{{{F_{rgb\_i} }}'\}_{i = 3}^5$. Then we add the complementary information to the another features respectively. And the another $1 \times 1$ 2D convolution is attached to learn more deeply for getting the enhanced paired features $\{{{F_{focal\_i} }}''\}_{i = 3}^5$ and $\{{{F_{focal\_i} }}''\}_{i = 3}^5$. The objective that using cross-modal residual connections to extract complementary characteristic from the paired features can be equivalently posed as approximating residual function. This reformulation disambiguates the multi-modal combination.

In the second step, inspired by \cite{meng2019frame}, we aggregate the enhanced RGB features and focal slices features with multiple attention weights. Take the enhanced ${F_{focal\_i}}''$ and ${F_{rgb_i}}''$ as an example, we concatenate them along the slice dimension and denote them as \emph{N} slices features $\{ {f_{i}^j } \}_{j = 1}^N$ (\emph{N}=13). First, in order to concentrate on depth information of every slice and the content information of RGB, we arrange coarse self-attention weights for every slice $f_{i}^j$. With these self-attention weights, we aggregate all the slices features into a global feature $F_{f_1}$. Because $F_{f_1}$ contains all focal slices features and RGB structure information over the entire depth range, we associate each slice features with the global feature representation $F_{f_1}$ to learn more reliable relation-attention weights for improving fusion results. As above, with the self-attention weights and associated attention weights, we integrate all the slice features to get the refine feature representation $F_{f_2}$. Finally, with a simple convolution layer, we can get the final fusion result $F_{i}$. The process can be defined as:
\begin{equation}
\begin{array}{l}
\gamma _j  = \sigma( {fc({dropout( {avgpooling(f_{i}^j )})})})
 \end{array}
\end{equation}
\begin{equation}
\begin{array}{l}
F_{f1}  = \frac{{\sum\nolimits_{j = 1}^N{\gamma_j f_{i}^j}}}{{\sum\nolimits_{j = 1}^N {\gamma _j }}}
 \end{array}
\end{equation}
\begin{equation}
\begin{array}{l}
\lambda _j  = \sigma ({fc({dropout({avgpooling({\rm{C}}({f_{i}^j ,F_{f1}})})})})
 \end{array}
\end{equation}
\begin{equation}
\begin{array}{l}
F_{f2}  = \frac{{\sum\nolimits_{j = 1}^N {\gamma _j \lambda _j {\rm{C}}( {f_{i}^j ,F_{f1} })} }}{{\sum\nolimits_{j = 1}^N {\gamma _j \lambda _j } }}
 \end{array}
\end{equation}
\begin{equation}
\begin{array}{l}
F_i = conv(F_{f2} ),
 \end{array}
\end{equation}
where $\sigma$ is the sigmoid function, $\gamma _j$ is the self-attention weight and $\lambda _j$ is the relation-attention weight of the \emph{j}-th slice features, \emph{C} is the concatenation operation. In summary, this module makes effective use of the complementarity between focal slices and the RGB image.

\section{Experiments}
\subsection{Dataset}
To evaluate the performance of our proposed network, we conduct experiments on two public light field datasets: DUT-LFDD \cite{piao2019depth}, LFSD dataset \cite{li2014saliency} and a mobile phone dataset \cite{suwajanakorn2015depth}.

\noindent {\bfseries{DUT-LFDD:}} This dataset contains 967 real-world light field samples which are captured by the Lytro camera. Each light filed consists of a RGB image, a focal stack with 12 focal slices focused at different depth and a corresponding ground truth depth map. We selects 967 samples which include 12 focal slices from the dataset. Specifically, we select 630 samples for training and the remaining 337 samples for testing. Previous studies in \cite{eigen2014depth,eigen2015predicting} show that data argumentation is helpful to improve accuracy as well as to avoid over-fitting. Therefore, to augment the training set, we flip the input images horizontally with $50\%$ chance, paying attention to swapping all images so they are in the correct position relative to each other. We also rotate them with a random degree in the ranges [-5, 5]. And we add color augmentations where we adopt random brightness, contrast, and saturation values by sampling from uniform distributions in the range [06, 1.4].

\noindent {\bfseries{LFSD:}} This dataset is proposed by Li \emph{et al}. It contains 100 light field scenes captured by Lytro camera, including 60 indoor and 40 outdoor scenes. Each scene contains an all-in-focus image, a focal stack with 12 focal slices and a depth map.

\noindent {\bfseries{Mobile Phone Dataset:}} This dataset is captured with a Samsung Galaxy phone during auto-focusing. Each scene is consist of a series of focal slices focused at different depth. It contains 13 scenes, such as plants, fruits, windows, etc. The size of every image is $640 \times 340$. 

\subsection{ Experiment setup}
\noindent {\bfseries{Evaluation Metrics.}} In order to comprehensively evaluate various methods, we adopt seven evaluation metrics commonly used in depth estimation task, including root mean squared error (rms),  mean absolute relative error (abs rel), squared relative error (sq rel), accuracy with a thread $\delta _i$. 

\noindent {\bfseries{Implementation details.}} Our method is implemented with pytorch toolbox and trained on a PC with GTX 2080 GPU. The input focal stack and RGB images are uniformly resized to $256 \times 256$. We use the ADM method for optimization with a initial learning rate 0.0001, and modify the learning rate to 0.00003 at 40 epochs. We initialize the backbone of our encoder with corresponding pre-trained VGG-16 net. The other layers in the network are randomly initialized. The batch size is 1 and maximum epoch is set 50. To improve prediction, we employ the loss function in \cite{hu2019revisiting} which consists of L1 loss, gradient loss and surface normal loss. We set the weights as 1 respectively in all the experiments.
\subsection{Ablation Studies}

In this section, we conduct ablation studies on each component of our network and further explore their functions.
\begin{figure}[!ht]
\begin{center}
\includegraphics[width=0.85\linewidth]{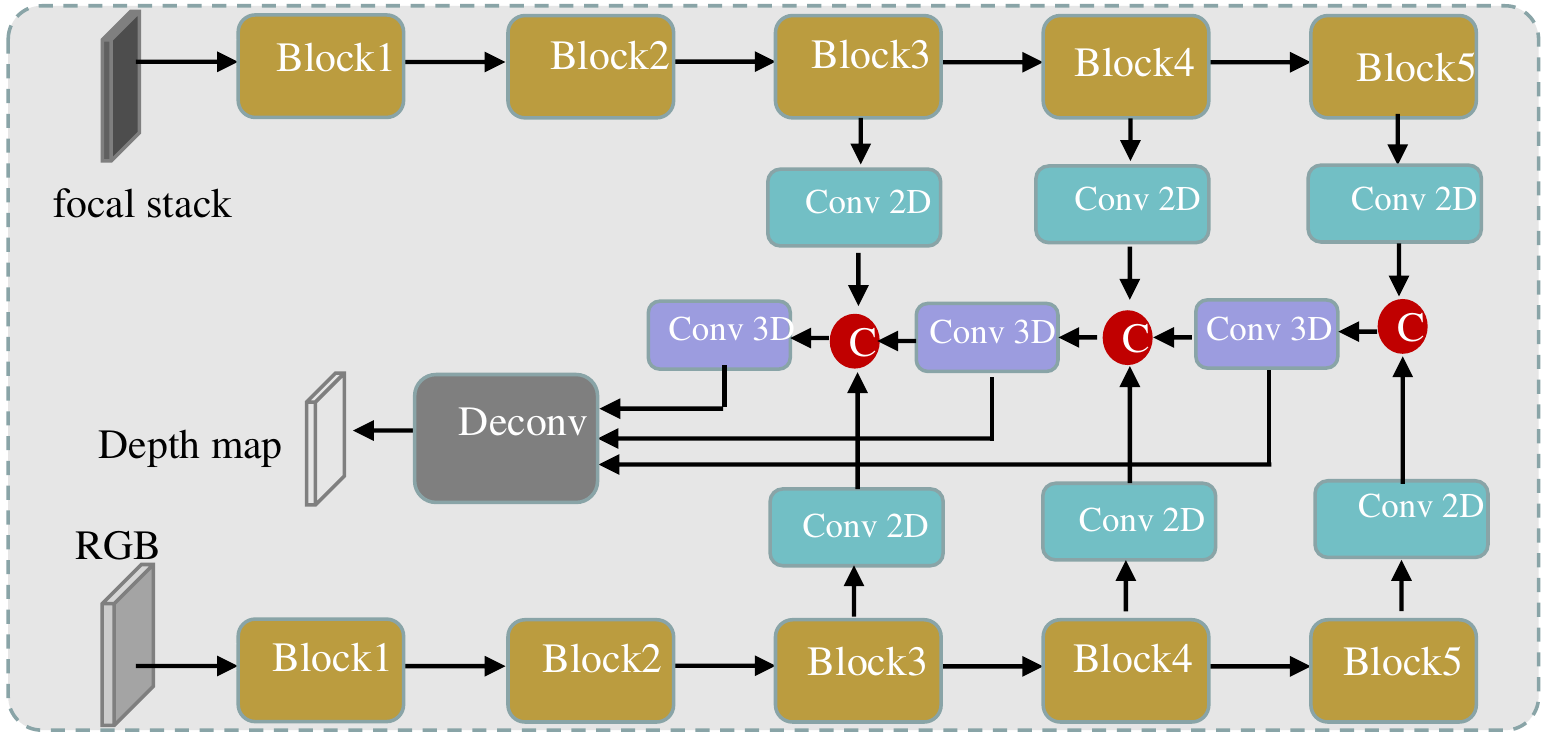}
\end{center}
\vspace{-4mm}
   \caption{The baseline network which is used to the RGB image and focal stack. }
\label{fig:short2}
\vspace{-6mm}
\end{figure}

\noindent {\bfseries{Baseline.}} For a fair comparison, we design a baseline network and denote it as 'Baseline'. As shown in Fig.3, the baseline network consists of two streams: focal stack stream and rgb stream. Each stream adopts the VGG-16 as the backbone. We also use Conv 2D and Conv 3D to replace CRU and CMFA in the proposed network, respectively. Each Conv 2D or Conv 3D contains 6 plain 2D convolution layers or 6 3D convolution layers. This fully ensures our baseline to have relatively high expressive power. We use the common concatenation to fuse the last three RGB features and corresponding focal slices features from the encoder network. Then we predict the depth map through the decoding network.

\noindent {\bfseries{Effect of the multi-modal input.}} To show the advantage of using multi-modal input, we provide the comparisons of using the RGB stream, focal stack stream and baseline. Specifically, the RGB stream and focal stack stream are achieved by corresponding streams in the baseline network, denoting as 'rgb' and 'focal stack' respectively. As shown in Table 1 and Table2, compared to the 'rgb' and 'focal stack', the 'Baseline' improve such high performance by a large margin. The depth maps in the Fig.4 also confirm that the combination of RGB images and the focal stack achieves better performance than using the single-model information alone.
\begin{table}[!ht]
  \centering
  \setlength{\tabcolsep}{2mm}
  \begin{threeparttable}
  \caption{Quantitative results of the ablation analysis on DUT-LFDD for our network. Note that $\delta_i = 1.25^{i} (i=1, 2,3)$.}
  \label{tab:performance_comparison}
    \begin{tabular}{ccp{1.05cm}<{\centering}p{0.9cm}<{\centering}p{0.8cm}<{\centering}p{0.8cm}<{\centering}p{0.8cm}<{\centering}p{0.8cm}<{\centering}p{0.8cm}<{\centering}}
    \toprule
    \multicolumn{1}{c}{\multirow{2}{*}{\scriptsize{Methods}}}&
    \multicolumn{4}{c}{\scriptsize error metric$\downarrow$}&\multicolumn{3}{c}{\scriptsize accuracy metric$\uparrow$}\cr
    \cmidrule(lr){2-4} \cmidrule(lr){5-7}
     &rms &rbs rel&sq rel&$\delta_1$&$\delta_2$&$\delta_3$\cr
     \midrule
     \multirow{1}{*} {\scriptsize {rgb }}&\scriptsize.4161&\scriptsize.1977&\scriptsize.1140&\scriptsize{.6520}&\scriptsize{.9164}&\scriptsize{.9867}\cr
     \multirow{1}{*} {\scriptsize {focal stack}}&\scriptsize.3856&\scriptsize.1830&\scriptsize.0978&\scriptsize{.6927}&\scriptsize{.9298}&\scriptsize{.9890}\cr
     \multirow{1}{*} {\scriptsize {Baseline}}&\scriptsize.3739&\scriptsize.1727&\scriptsize.0899&\scriptsize{.7020}&\scriptsize{.9373}&\scriptsize{.9919}\cr
    \multirow{1}{*} {\scriptsize {+CRU}}& \scriptsize{.3393}&\scriptsize{.1616}&\scriptsize.0793&\scriptsize{.7431}&\scriptsize{.9546}&\scriptsize{.9945}\cr
    \multirow{1}{*} {\scriptsize {+CMFA}}& \scriptsize{.3372}&\scriptsize{.1578}&\scriptsize.0767&\scriptsize{.7488}&\scriptsize{.9551}&\scriptsize{.9943}\cr
    \multirow{1}{*} {\scriptsize {+CRU(md)+CMFA}}&\scriptsize.3240&\scriptsize.1533&\scriptsize.0738&\scriptsize.7672&\scriptsize.9586&\scriptsize.9943\cr
    \multirow{1}{*} {\scriptsize
    {+CRU(mg)+CMFA}}&\scriptsize.3134&\scriptsize.1493&\scriptsize.0697&\scriptsize.7757&\scriptsize.9644&\scriptsize.9945\cr
    \multirow{1}{*} {\scriptsize
    {+CRU+CMFA(Ours)}}&\scriptsize.3029&\scriptsize.1455&\scriptsize.0668&\scriptsize.7859&\scriptsize.9685&\scriptsize.9956\cr
    \bottomrule
    \end{tabular}
    \end{threeparttable}
\end{table}
\begin{table}[!ht]
 \centering
  \setlength{\tabcolsep}{2mm}
  \begin{threeparttable}
  \caption{Quantitative results of the ablation analysis on LFSD for our network.}
  \label{tab:performance_comparison}
    \begin{tabular}{ccp{1.05cm}<{\centering}p{0.9cm}<{\centering}p{0.8cm}<{\centering}p{0.8cm}<{\centering}p{0.8cm}<{\centering}p{0.8cm}<{\centering}p{0.8cm}<{\centering}}
    \toprule
    \multicolumn{1}{c}{\multirow{2}{*}{\scriptsize{Methods}}}&
    \multicolumn{4}{c}{\scriptsize error metric$\downarrow$}&\multicolumn{3}{c}{\scriptsize accuracy metric$\uparrow$}\cr
    \cmidrule(lr){2-4} \cmidrule(lr){5-7}
     &rms&rbs rel&sq rel&$\delta_1$&$\delta_2$&$\delta_3$\cr
     \midrule
     \multirow{1}{*} {\scriptsize {rgb}}&\scriptsize.4637&\scriptsize.2098&\scriptsize.1286&\scriptsize{.6013}&\scriptsize{.8855}&\scriptsize{.9797}\cr
     \multirow{1}{*} {\scriptsize {focal stack}}&\scriptsize.4106&\scriptsize.1821&\scriptsize.1000&\scriptsize{.6757}&\scriptsize{.9198}&\scriptsize{.9885}\cr
     \multirow{1}{*} {\scriptsize {Baseline}}&\scriptsize.4029&\scriptsize.1791&\scriptsize.0957&\scriptsize{.6785}&\scriptsize{.9327}&\scriptsize{.9899}\cr
    \multirow{1}{*} {\scriptsize {+CRU}}&
    \scriptsize{.3727}&\scriptsize{.1660}&\scriptsize.0833&\scriptsize{.7136}&\scriptsize{.9420}&\scriptsize{.9951}\cr
    \multirow{1}{*} {\scriptsize {+CMFA}}& \scriptsize{.3647}&\scriptsize{.1622}&\scriptsize.0807&\scriptsize{.7257}&\scriptsize{.9412}&\scriptsize{.9937}\cr
     \multirow{1}{*} {\scriptsize {+CRU(md)+CMFA}}&\scriptsize.3503&\scriptsize.1566&\scriptsize.0753&\scriptsize.7513&\scriptsize.9546&\scriptsize.9947\cr
    \multirow{1}{*} {\scriptsize {+CRU(mg)+CMFA}}&\scriptsize.3316&\scriptsize.1490&\scriptsize.0669&\scriptsize.7718&\scriptsize.9658&\scriptsize.9964\cr
    \multirow{1}{*} {\scriptsize {+CRU+CMFA(Ours)}}&\scriptsize.3167&\scriptsize.1426&\scriptsize.0627&\scriptsize.7814&\scriptsize.9686&\scriptsize.9964\cr
    \bottomrule
    \end{tabular}
    \end{threeparttable}
\end{table}

\begin{figure}[!ht]
\begin{center}
\includegraphics[width=0.9\linewidth]{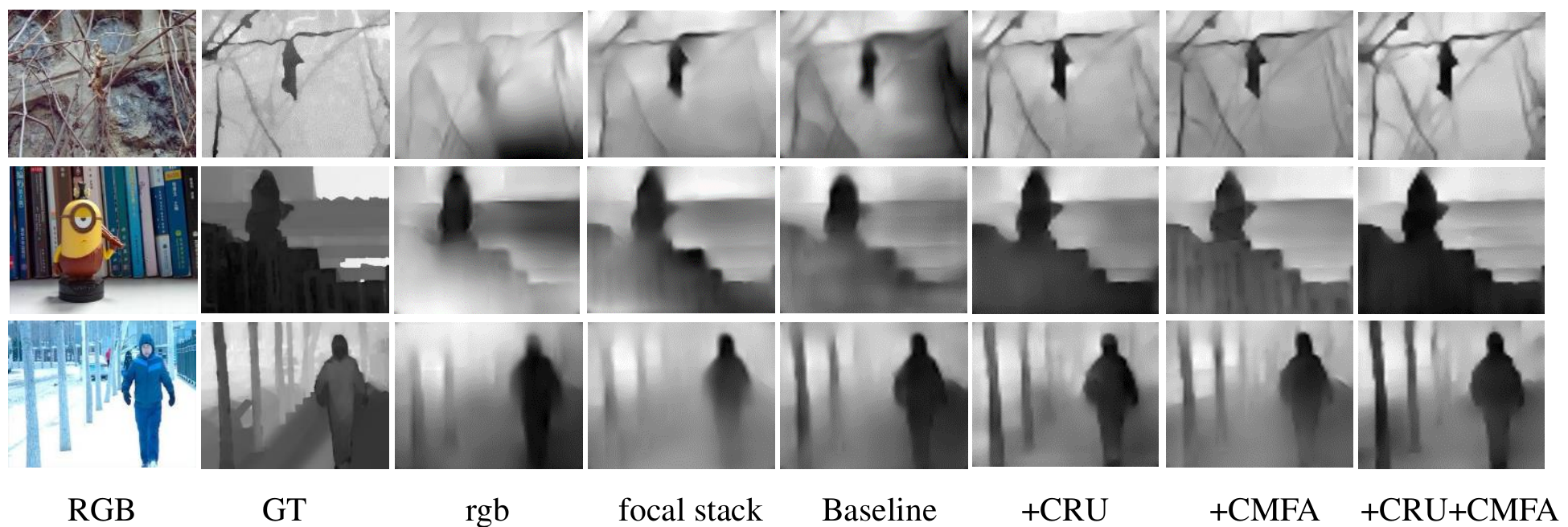}
\end{center}
\vspace{-4mm}
   \caption{The visual results of ablation analysis.}
\label{fig:short2}
\vspace{-6mm}
\end{figure}

\noindent {\bfseries{Effect of the Context Reasoning Unit (CRU).}} The CRU is proposed to comprehensively extract contextual information from the focal slices and RGB features for exploring the internal spatial correlation. It effectively reasons the relationship between focus and defocus region of focal slices and the comprehensive structure relation between different regions of RGB images. In order to verify the effectiveness of CRU, we use CRU to replace the Conv 2D of baseline network and evaluate its performance (denoted as '+CRU'). As shown in Table 1 and Table 2, '+CRU' significantly outperforms 'Baseline' on all evaluation indicators. Compared to the baseline, as shown in $1^{st}$ row in the Fig.4, with the help of the CRU, the depth change on the small and thin objects is more obvious. Its powerful relational reasoning ability can better excavate correlations between objects and different regions within the image to clarify the local depth confusion.

\noindent {\bfseries{Effect of the Attention-guide Cross-Modal Fusion module (CMFA).}} The CMFA is proposed to integrate the rich information in focal slices and RGB images to compensate for detail loss caused by defocus blur. In order to prove that our fusion method can capture complementary features more easily than simple concatenation, We use the CMFA model (noted as '+CMFA') to replace the simple concatenation and violent 3D convolution layers in 'Baseline' between the corresponding hierarchical features. The quantitative results in Table 1 and Table 2 and visual results in Fig.4 both show that our '+CMFA' can better fuse the information of the paired focal slices and RGB features than baseline network. It greatly reduces the estimation error over the dataset and achieves impressive accuracy improvements.

\noindent {\bfseries{Effect of the CRU and CMFA.}} In order to prove that our reasoning unit and fusion method can jointly extract and aggregate the different information in the focal stack and the RGB image, we achieve our final methods by using them together. Moreover, we specifically explore the effectiveness of every component in the CRU. For convenience, we denote 'the multiple dilated convolutions' and 'the multiple graph convolutions' as 'CRU (md)' and 'CRU (mg)', respectively.

Compared to our '+CRU' and '+CMFA', combining the 'CRU (md)' module with 'CMFA' (noted as '+CRU(md)+CMFA') and combining the 'CRU (mg)' module with 'CMFA' (noted as '+CRU(mg)+CMFA') both obtains an improvement by a obvious margin. The best results are obtained by combining the two modules together, denoted as '+CRU+CMFA'. As shown in Table 1 and Table 2, compared to 'Baseline', the RMSE value is reduced by nearly ${7\%}$ on DUT-LFDD and ${9\%}$ on LFSD. Our '+CRU+CMFA' greatly reduced the estimation error over the entire dataset and achieve impressive accuracy improvements. This shows our modules can effectively achieve the respective functions and do not interfere with each other. From the visual results in Fig.4, we can clearly observe that the depth maps of '+CRU+CMFA' have more complete information. It obviously prove that our method can better refine and fuse the focal slices features and RGB features.

\subsection{Comparisons with State-of-the-arts}

We compare results from our method and other six state-of-arts methods, containing both deep-learning-based methods(\emph{DDFF} \cite{hazirbas2018deep}, \emph{EPINet} \cite{shin2018epinet}) and non-deep learning method marked with * (\emph{PADMM}$^*$ \cite{javidnia2018application}, \emph{VDFF}$^*$ \cite{moeller2015variational}, \emph{LFACC}$^*$  \cite{jeon2015accurate}, \emph{LF}$_-$\emph{OCC}$^*$ \cite{wang2015occlusion}). For fair comparisons, we use the parameter settings provided by authors and adjust some of the parameters to fit different datasets as needed. Note that because the LFSD
dataset does not contain multi-view images, results of some methods are not available.

\begin{table*}[!ht]
  \centering
  \setlength{\tabcolsep}{1mm}
  \begin{threeparttable}
  \caption{Quantitative comparisons with state-of-the-art methods. From top to bottom: DUT-LFDD Dataset, LFSD dataset. $*$ respects non-deep-learning methods. The best results are shown in \textbf{boldface}.}
  \label{tab:performance_comparison}
    \begin{tabular}{ccp{0.8cm}<{\centering}p{1.05cm}<{\centering}p{0.9cm}<{\centering}p{0.8cm}<{\centering}p{0.8cm}<{\centering}p{0.8cm}<{\centering}p{0.8cm}<{\centering}}
    \toprule
    \multicolumn{1}{c}{\multirow{2}{*}{type}}&
    \multicolumn{1}{c}{\multirow{2}{*}{methods}}&
    \multicolumn{4}{c}{error metric}&\multicolumn{3}{c}{accuracy metric}\cr
    \cmidrule(lr){3-5} \cmidrule(lr){6-8}
     &{}&rms&rbs rel&sq rel&$\delta_1$&$\delta_2$&$\delta_3$\cr
    \midrule
     \multirow{7}{*}{DUT-LFDD}&Ours&{\bfseries{.3029}}&{\bfseries{.1455}}&{\bfseries{.0668}}&{\bfseries{.7859}}&{\bfseries{.9685}}&{\bfseries{.9956}}\cr
    &DDFF&.5282&.2666&.1838&.4817&.8196&.9658\cr
    &EPINet&.4974&.2324&.1434&.5010 &.8375 &.9837\cr
    &VDFF$^*$&.7326&.3689&.3303&0.3348&.6283&.8407\cr
    &PADMM$^*$&.4730&.2253&.1509&.5891&.8560&.9577\cr
    &LFACC$^*$&.6897&.3835&.3790&.4913&.7549&.8783\cr
    &LF$_-$OCC$^*$&.6233&.3109&.2510&.4524&.7464&.9127\cr
    \midrule
     \multirow{8}{*}{LFSD}&Ours&{\bfseries{.3167}}&{\bfseries{.1426}}&{\bfseries{.0627}}&{\bfseries{.7814}}&{\bfseries{.9686}}&{\bfseries{.9964}}\cr
    &DDFF&.6222&.3593&.2599&.3447&.7352&.9476\cr
    &EPINet&-&-&-&-&-&- \cr
    &VDFF$^*$&.7842&.4736&.5076&.3631&.5932&.8150\cr
    &PADMM$^*$&.3395&.1798&.1012&.7661&.9363&.9744\cr
    &LFACC$^*$&-&-&-&-&-&-\cr
    &LF$_-$OCC$^*$&-&-&-&-&-&-\cr
    \midrule
    \end{tabular}
    \end{threeparttable}
\vspace{-2mm}
\end{table*}

\begin{figure*}[!ht]
\begin{center}
\includegraphics[width=0.9\linewidth]{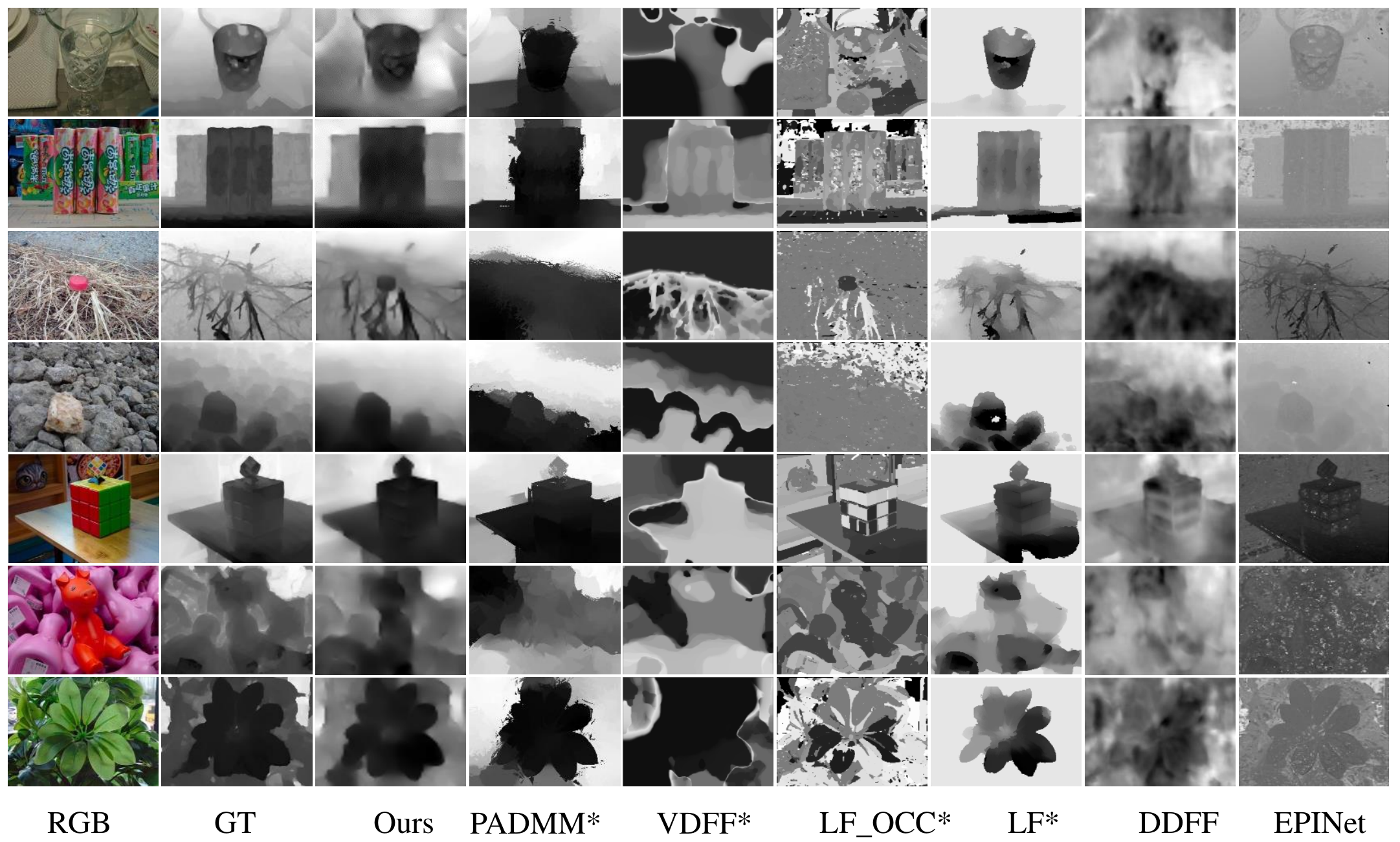}
\end{center}
\vspace{-4mm}
   \caption{ Comparisons with the other methods on DUT-LFDD dataset: RGB images, the corresponding ground truth,  estimated results of different methods.}
\label{fig:long}
\vspace{-2mm}
\end{figure*}

\begin{figure}[!ht]
\begin{center}
\includegraphics[width=0.9\linewidth]{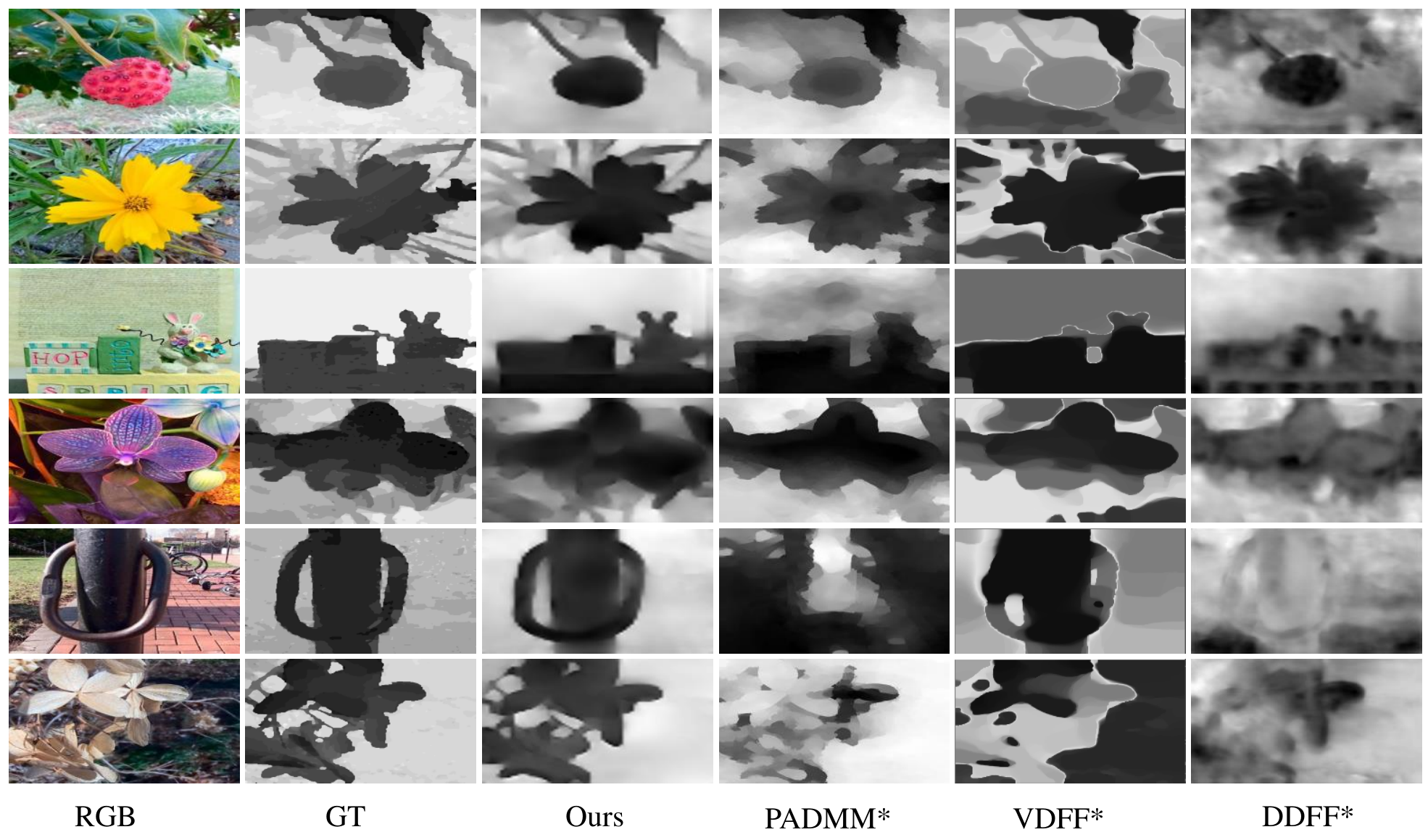}
\end{center}
\vspace{-4mm}
   \caption{The visual results of our method on the LFSD dataset.}
\label{fig:short1}
\vspace{-4mm}
\end{figure}

\noindent {\bfseries{Quantitative Evaluation.}} As shown in Table 3, compared to other methods, our network can achieve significant superior performance in terms of all evaluation metrics on the LFSD dataset and DUT-LFDD dataset. It indicates that our model is more powerful. More specifically, due to the limited generalization ability brought by the reliance on prior knowledge, all non-deep-learning methods are unable to get better results on both datesets. Compared to DDFF, our method successfully introduces the RGB information, which significantly enhances the depth map. Moreover, benefit from more sufficient contextual information reasoning and fusion of multi-modal features, our method still achieves much better performance than the DDFF and EPINet.

\noindent {\bfseries{Qualitative Evaluation. }}We also visually compare our method with the representative method on the DUT-LFDD and LFSD datasets. As shown in Fig.5 and Fig.6, our method is able to handle a wide rage of challenging scenes. In those challenging cases, such as the similar foreground and background, multiple or transparent objects and complex background, our method can highlight depth change and fine detail information. In contrast, most of other methods are unlikely to predict more correct depth value due to the lack of high-level contextual reasoning or robust multi-modal fusion strategy. However, the proposed network is able to utilize both cross-modal and cross-level complementary information to learn cooperatively discriminative depth cues. Although the boundary of the resultant depth map seems a bit blurry, it is more accurate and reduces much of the noise. These visual results powerfully verify our proposed network are more effective and robust.

\begin{figure}[!ht]
\begin{center}
\includegraphics[width=0.9\linewidth]{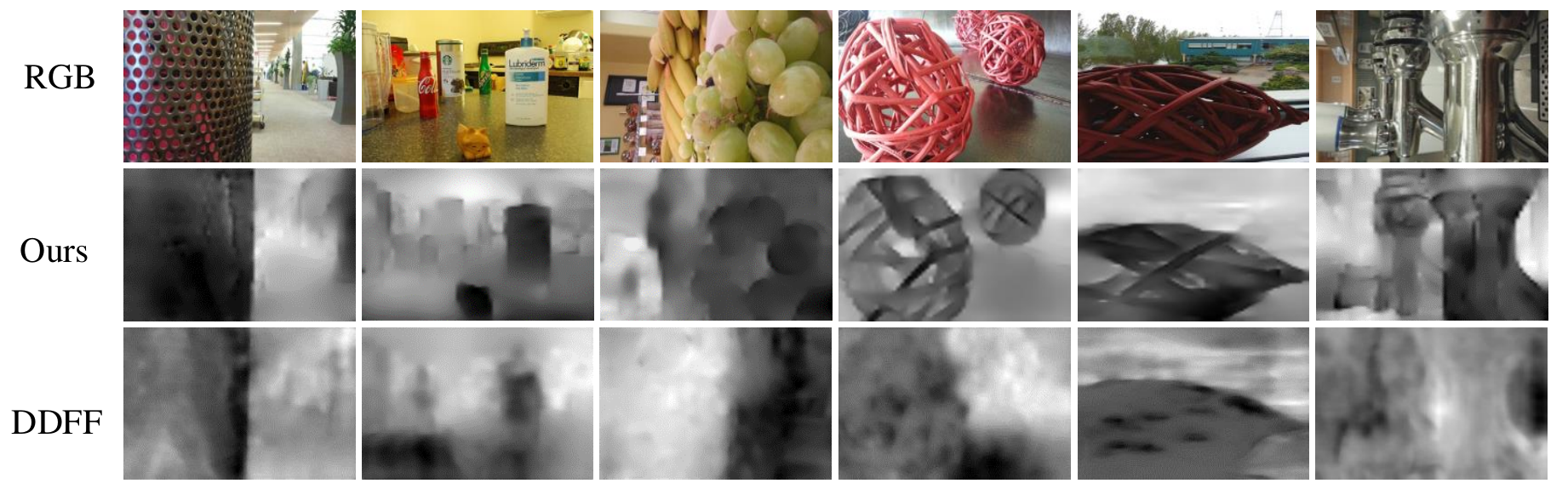}
\end{center}
\vspace{-2mm}
   \caption{The visual results of our method on the mobile phone dataset.}
\label{fig:short1}
\vspace{-4mm}
\end{figure}

\noindent {\bfseries{Adaptation on the Mobile phone Dataset.}} To demonstrate the applicability of our method, we evaluate our framework on the smart-phone camera dataset \cite{suwajanakorn2015depth} which are pre-aligned. Note that alignment should
be considered for practical applications. We feed 12 focal slices and RGB image from this dataset into our framework. As shown in Fig.7, our method can be generalized to the mobile phone dataset easily. Compared with the results from DDFF, our method captures more detail information. These demonstrate our method is more suitable for daily life.

\section{Conclusions}
In this paper, we propose a effective network for predicting depth maps from the focal stack and the RGB image, which can learn end-to-end. Our method enhances the performance from the following aspects: 1) comprehensively extract the contextual information reasons to explore internal spatial correlation by using a effective context reasoning module (CRU); 2) effectively fuses paired contextual information extracted from the focal stack and the RGB image by attention-guided cross-modal fusion module (CMFA). We thoroughly validate the effectiveness of each component in the network and gradually show an increase in cumulative accuracy. Experiment results also demonstrate that our method achieves new state-of-the-art performance on two light field datasets and a mobile phone dataset.

%
%
%
%
\bibliographystyle{splncs04}
\bibliography{egbib}
\end{document}